\documentclass{article}               
\usepackage{natbib}
\usepackage[hidelinks]{hyperref}
\usepackage{graphicx}
\usepackage{xcolor}
\usepackage[margin=3.2cm]{geometry}

\begin{document}

\title{AI2D-RST: A multimodal corpus of 1000 primary school science diagrams}

\author{Tuomo Hiippala \and Malihe Alikhani \and Jonas Haverinen \and Timo Kalliokoski \and Evanfiya Logacheva \and Serafina Orekhova \and Aino Tuomainen \and Matthew Stone \and John A. Bateman}

\date{}

\maketitle

\begin{abstract}
This article introduces AI2D-RST, a multimodal corpus of 1000 English-language diagrams that represent topics in primary school natural sciences, such as food webs, life cycles, moon phases and human physiology. The corpus is based on the Allen Institute for Artificial Intelligence Diagrams (AI2D) dataset, a collection of diagrams with crowd-sourced descriptions, which was originally developed to support research on automatic diagram understanding and visual question answering. Building on the segmentation of diagram layouts in AI2D, the AI2D-RST corpus presents a new multi-layer annotation schema that provides a rich description of their multimodal structure. Annotated by trained experts, the layers describe (1) the grouping of diagram elements into perceptual units, (2) the connections set up by diagrammatic elements such as arrows and lines, and (3) the discourse relations between diagram elements, which are described using Rhetorical Structure Theory (RST). Each annotation layer in AI2D-RST is represented using a graph. The corpus is freely available for research and teaching.
\end{abstract}

\section{Introduction}

Diagrams are a common feature of many everyday media: they can be found everywhere from scientific publications and instruction manuals to newspapers and school textbooks. Barbara Tversky, a cognitive psychologist who has made pioneering contributions to the study of diagrams, observes that their generic purpose is ``to structure information to enable comprehension, inference and discovery'' \citep[350]{tversky2017}. Due to their widespread use, diagrams have been studied from various perspectives. Previous research has examined their visual perception \citep{hegartyjust1993, ware2012}, structure and functions \citep{engelhardt2002, purchase2014, engelhardtrichards2018} and their role as a tool for supporting thinking and reasoning \citep{tversky2015} and use in education and instruction \citep{tippett2016}, to name but a few examples.

In this article, we make a novel contribution to the study of diagrams by presenting AI2D-RST, a corpus of 1000 English-language diagrams that represent topics in primary school natural sciences. The diagrams are described using a new multi-layer annotation schema that seeks to capture their multimodal structure. Our approach to multimodality is linguistically-inspired and semiotically-oriented, that is, we seek to systematically describe how expressive resources such as natural language, illustrations, line art, photographs, lines, arrows and layout are combined in diagrams to make and exchange meanings. To do so, we build on the general framework for multimodal communication proposed in \citet{batemanetal2017} and its application to diagrams as set out in \citet{hiippalabateman2020}.

The current work is situated within the emerging field of multimodality research, which studies how appropriate combinations of expressive resources emerge in communicative situations \citep[see e.g.][]{wildfeueretal2020}. Despite their growing influence in various fields of study broadly concerned with human communication, many approaches to multimodality remain without adequate empirical support. Although building multimodal corpora is often presented as a solution to this shortcoming due to the success of corpus-based methods in linguistics, developing and applying complex multimodal annotation frameworks requires ample time and resources, and consequently the resulting corpora remain small \citep{bateman2008, bateman2014, thomas2014, hiippala2015a}.

AI2D-RST seeks to reduce the need for time and resources and to scale up the volume of data by building multimodally-informed expert annotations on top of pre-existing crowd-sourced annotations from another dataset, namely the Allen Institute for Artificial Intelligence Diagrams (AI2D) dataset \citep{kembhavietal2016}. The second part of the name, RST, refers to Rhetorical Structure Theory, a theory of discourse structure which we use to describe how diagrams combine multiple expressive resources to fulfil their communicative goals \citep{mannthompson1988, taboadamann2006a, hiippalaorekhova2018}. Overall, the AI2D-RST corpus is intended to serve a dual purpose: to support empirical research on the multimodality of diagrams and their computational processing.

\section{Developing multimodal resources for diagrams research}
\label{sec:resources}

There is a long-standing interest in the computational processing and generation of diagrammatic representations \citep{andrerist1995,watanabenagao1998,batemanetal2001,carberryetal2003,batemanhenschel2007}, which is now resurfacing as recent advances in computer vision and natural language processing are brought to bear on diagrammatic representations \citep{seoetal2015,sachanetal2018,sachanetal2019,choietal2018,kimetal2019,haehnetal2019}. Much of this work is driven by research on established computational tasks such as information retrieval and question answering, which now increasingly acknowledge that the scope of these tasks should be extended to modes of expression beyond natural language.

Just how these other modes of expression \emph{and} their combinations should be described in order to create multimodal resources that can support further research on multimodality remains an open question. This requires an empirical approach, as creating multimodal resources for modes of expression other than natural language raises questions about fundamental issues such as segmentation: how to decompose modes of expression such as diagrams into their constituent parts? We have recently argued in \citet{hiippalabateman2020} that any attempt at a systematic description of diagrams must acknowledge the specific characteristics of the \emph{diagrammatic mode} -- an abstract system capable of instantiating various types of diagrams appropriate for their context of occurrence \citep[cf. e.g.][]{batemanhenschel2007}.

Previous research points at two key characteristics of the diagrammatic mode that need to be accounted for: the use of layout space \citep{watanabenagao1998} and their multimodal discourse structure \citep{carberryetal2003}, which are often strongly intertwined in multimodal artefacts with a 2D spatial extent, such as entire page-based documents \citep{hiippala2013a}. Firstly, diagrams have a \emph{spatial} organisation in the form of a layout, which can be used to set up discourse relations between instances of expressive resources, including natural language, arrows, lines, illustrations, photographs, line drawings and potentially any resource that may be realised in 2D space \citep{watanabenagao1998}. How expressive resources are organized in the layout space can also serve as a strong signal about the purpose and structure of the diagram by generating expectations towards its discourse structure \citep{holsanovaetal2009}.

This brings us to the second point: diagrams combine expressive resources into \emph{discourse structures}, which must be resolved to make sense of what the diagram in question attempts to communicate. For this reason, \citet{carberryetal2003} argue that understanding diagrams should be framed a discourse-level problem, which has found support in our recent work on the diagrammatic mode \citep{hiippalabateman2020}. This, however, raises another issue related to segmentation: many theories of discourse assume that discourse segments are identified before determining their interrelations \citep{groszsidner1986,mannthompson1988}.

Establishing an inventory of discourse segments for diagrams is a particularly challenging task, as the level of detail needed for segmentation varies from one diagram to another, depending on the combination of expressive resources present and the discourse structures they participate in. To exemplify, a diagram featuring a cross-section of an object may need to be decomposed to parts that are picked out by labels, whereas illustrations that feature entire objects do not require such treatment \citep[for a discussion of segmentation, see][]{hiippalabateman2020}.

Keeping these two key characteristics of diagrams in mind, in the following sections we explicate how we built a new, multimodally-informed annotation schema with multiple layers of description on top of existing crowd-sourced annotations for expressive resources and their positioning in diagram layout. To do so, we start by introducing the AI2D dataset, which provided the crowd-sourced annotations. We then address certain issues with the AI2D annotation schema before motivating our decision to adopt Rhetorical Structure Theory for describing the discourse structure of diagrams in AI2D-RST.

\section{The Allen Institute for Artificial Intelligence Diagrams (AI2D) dataset}
\label{sec:ai2d}

The AI2D dataset \citep{kembhavietal2016}\footnote{The AI2D dataset is publicly available at \url{https://allenai.org/plato/diagram-understanding/} (Accessed September 3, 2019)} was developed to support research on computational tasks such as automatic diagram understanding and visual question answering \citep[see e.g.][]{kimetal2018}. The dataset contains 4903 English-language diagrams that represent topics in primary school natural sciences, such as life cycles, food webs and circuits. Each diagram is assigned to one of 17 semantic categories that correspond to topics in this domain.

\begin{figure}[h!]
    \centering
    \includegraphics[width=1\textwidth]{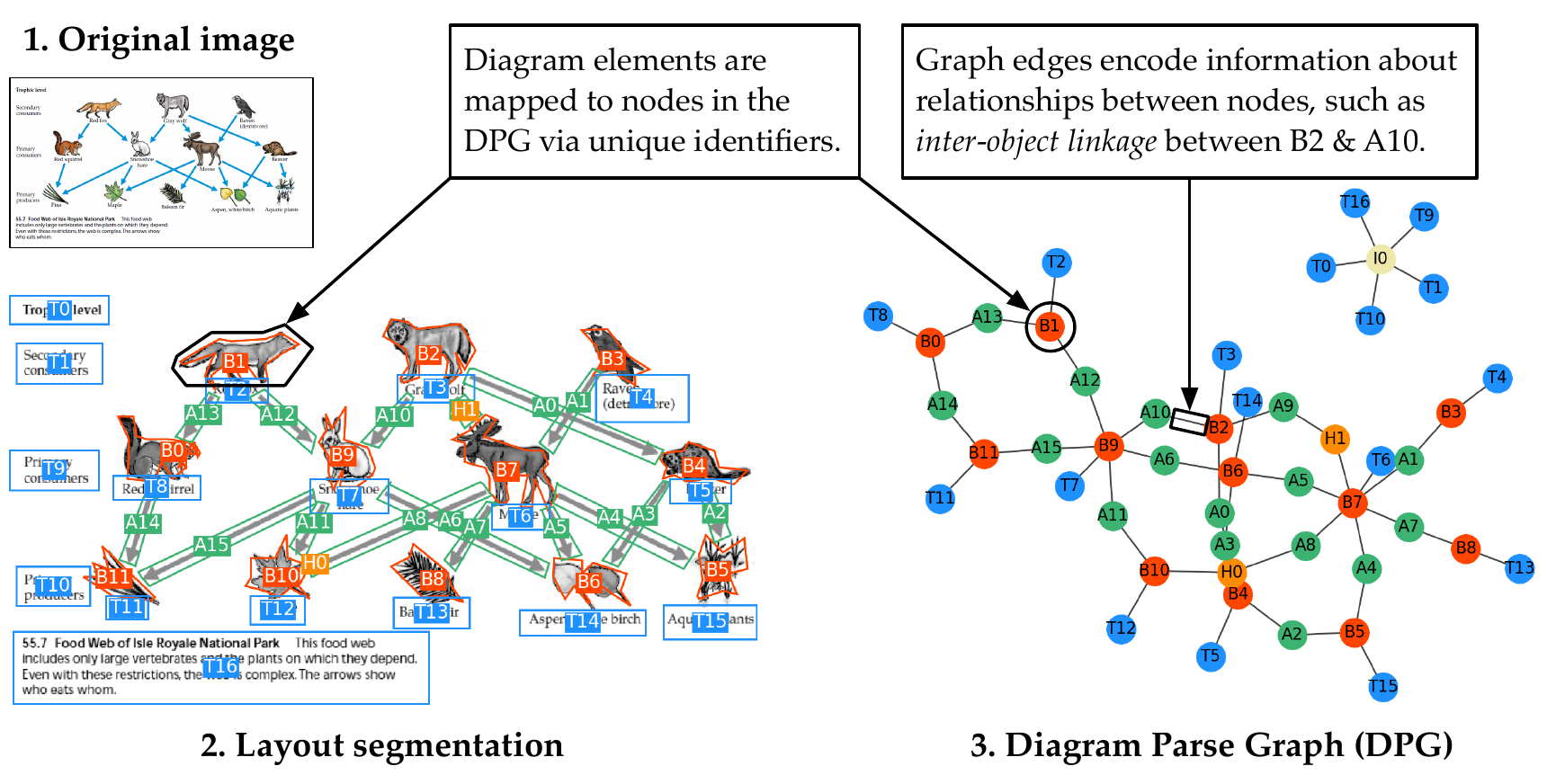}
    \caption{(1) A thumbnail of the original diagram image scraped from the web, (2) its crowd-sourced layout segmentation (converted into greyscale to bring out the annotation) and (3) a Diagram Parse Graph (DPG) for diagram \#274 in AI2D. Diagram element types are coded using same colours in both layout segmentation and DPG: text blocks (blue), blobs (red), arrows (green), arrowheads (orange) and image constant (Navajo white).}
    \label{fig:dpg}
\end{figure}

Building on \citeauthor{engelhardt2002}'s \citeyearpar{engelhardt2002} framework for describing diagrammatic representations, \citet[239]{kembhavietal2016} model four types of diagram elements: `blobs' (e.g. illustrations, line art, photographs and other visual modes of expression), written text, arrows and arrowheads. In addition, \citet{kembhavietal2016} define ten potential relationships that can hold between individual diagram elements, which are also drawn from the framework proposed by \citet{engelhardt2002}. These include, among others, relations such as \textsc{intra-object label}, \textsc{intra-object linkage} and \textsc{arrow descriptor}, which seek to capture how diagram elements relate to each other \citep[for a full list of relations, see][239]{kembhavietal2016} AI2D represents diagram structure using a \emph{Diagram Parse Graph} (DPG), in which the nodes stand for diagram elements whereas the edges encode information about the relations that hold between them. For computational tasks, the node features can be populated using word embeddings or visual features extracted using object detectors, depending on the diagram element type in question.

Figure \ref{fig:dpg} exemplifies the crowd-sourced annotation for the layout segmentation and DPG for diagram 274 in the AI2D dataset. The diagrams were scraped from Google Image Search by using chapter titles in primary school science textbooks (for ages 6--11) as search terms. The annotations were crowd-sourced using Amazon Mechanical Turk and divided into small tasks to segment the layout and construct a DPG for each diagram. The annotation tasks involved identifying diagram elements, categorising them and defining their interrelations \citep[243]{kembhavietal2016}. \citet[242]{kembhavietal2016} report that the 4903 diagrams in AI2D contain approximately 118 000 diagram elements and 53 000 relationships.

Previous research on the AI2D dataset has shown that inferring the meaning of arrows and lines is context-dependent, and the viewers consistently map the arrows to real-world processes they represent \citep{alikhanistone2018}. \citet{hiippalaorekhova2018}, in turn, consider AI2D from the perspective of multimodality research and argue that DPGs conflate the description of various multimodal structures, such as the visual grouping of diagram elements and connections expressed using arrows and lines. Pulling these structures apart could help to understand how diagrams operate multimodally, that is, whether discourse relations are signalled explicitly or implicitly \citep{watanabenagao1998, carberryetal2003}.

\begin{figure}[h!]
    \centering
    \includegraphics[width=0.49\textwidth]{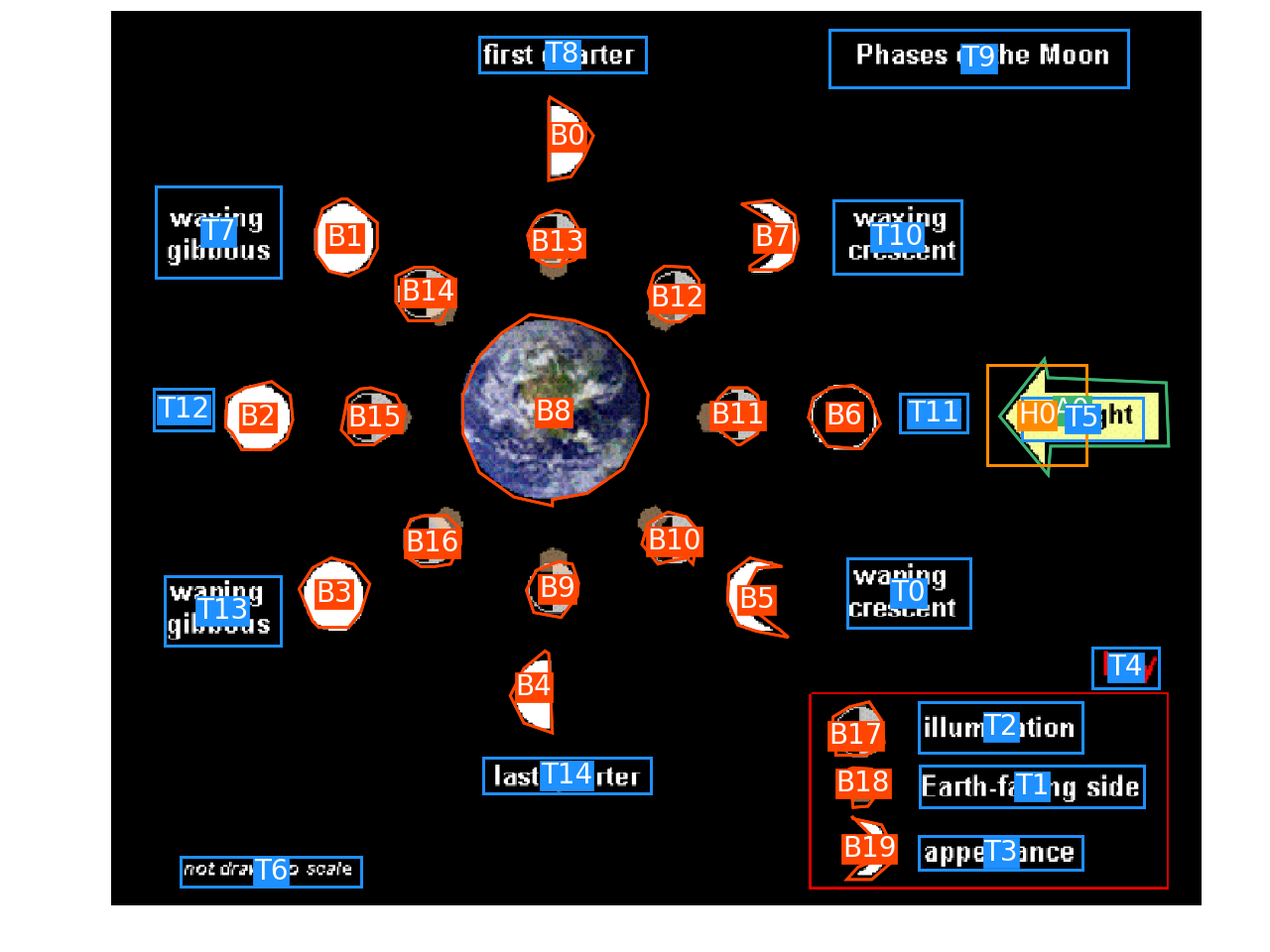} \includegraphics[width=0.49\textwidth]{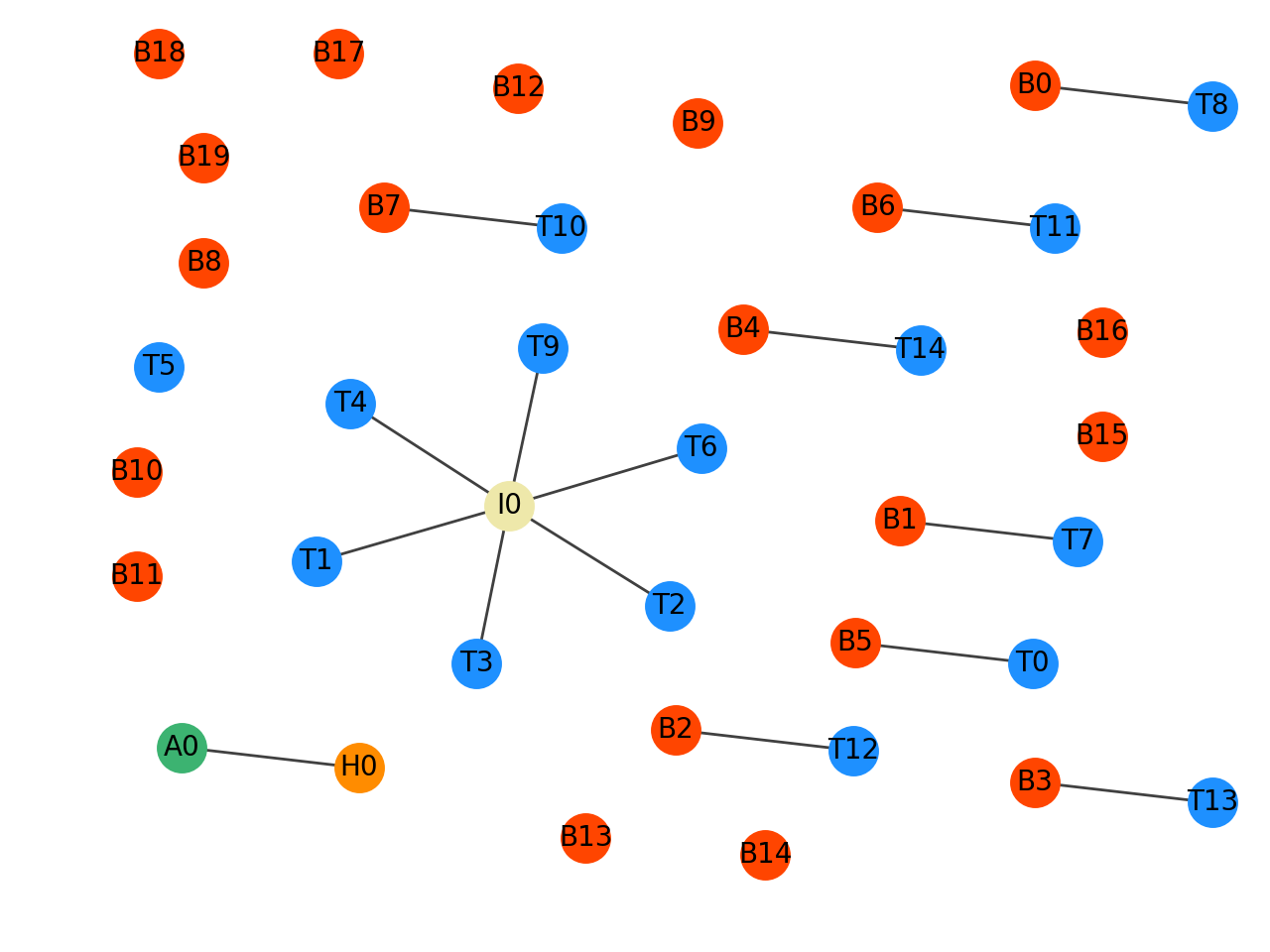}
    \caption{Layout segmentation (left) and Diagram Parse Graph (DPG, right) for diagram \#2728. The numerous disconnections in the DPG result from the lack of relation definitions for describing how \emph{groups} of diagram elements, such as those formed by illustrations of moon phases and their verbal descriptions (e.g. illustration B3 and the text `waning gibbous' in T13), relate to each other as a part of the global discourse structure of the diagram.}
    \label{fig:problem}
\end{figure}

Moreover, the relation definitions drawn from \citet{engelhardt2002} focus mainly on \emph{local} relations between diagram elements, as exemplified by relations such as \textsc{intra-object label}, which is used to describe instances in which one diagram element acts as a label for another. The focus on such local relations between individual diagram elements causes the AI2D annotation schema to fall short in describing the \emph{global} organisation of a diagram, or how larger units formed of multiple diagram elements relate to each other (see Figure \ref{fig:problem}).

To summarize, the motivation for developing AI2D-RST can be traced back to two observations. First, the limited scope of relation definitions drawn from \citet{engelhardt2002} in AI2D led us to consider Rhetorical Structure Theory (RST) as an alternative for describing discourse relations in diagrams, given its previous successful applications to multimodal discourse \citep[see e.g.][]{taboadahabel2013, thomas2014, hiippala2015a}. However, during the exploratory work reported in \citet{hiippalaorekhova2018}, it became evident that a direct conversion to RST was not feasible, but required defining additional annotation layers to establish the units of analysis, as proposed in \citet{bateman2008}.

Second, combining a theory of discourse structure with local and global reach, such as RST, with a multi-layer annotation schema that captures the combinations of expressive resources and their spatial organisation could be used to study whether diagrams signal discourse relations explicitly e.g. using arrows and lines, or whether they are implicit and require the viewers to draw on world knowledge \citep[see also][]{hiippalabateman2020}. Furthermore, access to crowd-sourced layout segmentations allows scaling up corpus size. With these two observations in mind, we now turn to describe the AI2D-RST annotation schema and its application to the AI2D diagrams.

\section{Developing the AI2D-RST corpus}
\label{sec:annotation}

\subsection{The AI2D-RST annotation schema}

The AI2D-RST annotation schema describes the multimodal structure of diagrams using four annotation layers. These layers, named \emph{grouping}, \emph{macro-grouping}, \emph{connectivity} and \emph{discourse structure}, are introduced in the following sections. The annotation layers are represented using graphs, which are populated using diagram elements from the AI2D layout segmentation (see Figure \ref{fig:dpg}). The unique identifiers for diagram elements are also carried over from the AI2D layout segmentation to the AI2D-RST graphs, in order to enable cross-references across annotation layers. This kind of stand-off approach to annotation separates the description of different multimodal structures, but allows combining them as necessary using the unique identifiers, which are shared across annotation layers.

\subsubsection{Grouping} 
\label{sec:grouping}

\begin{figure}[h!]
    \centering
    \includegraphics[width=1\textwidth]{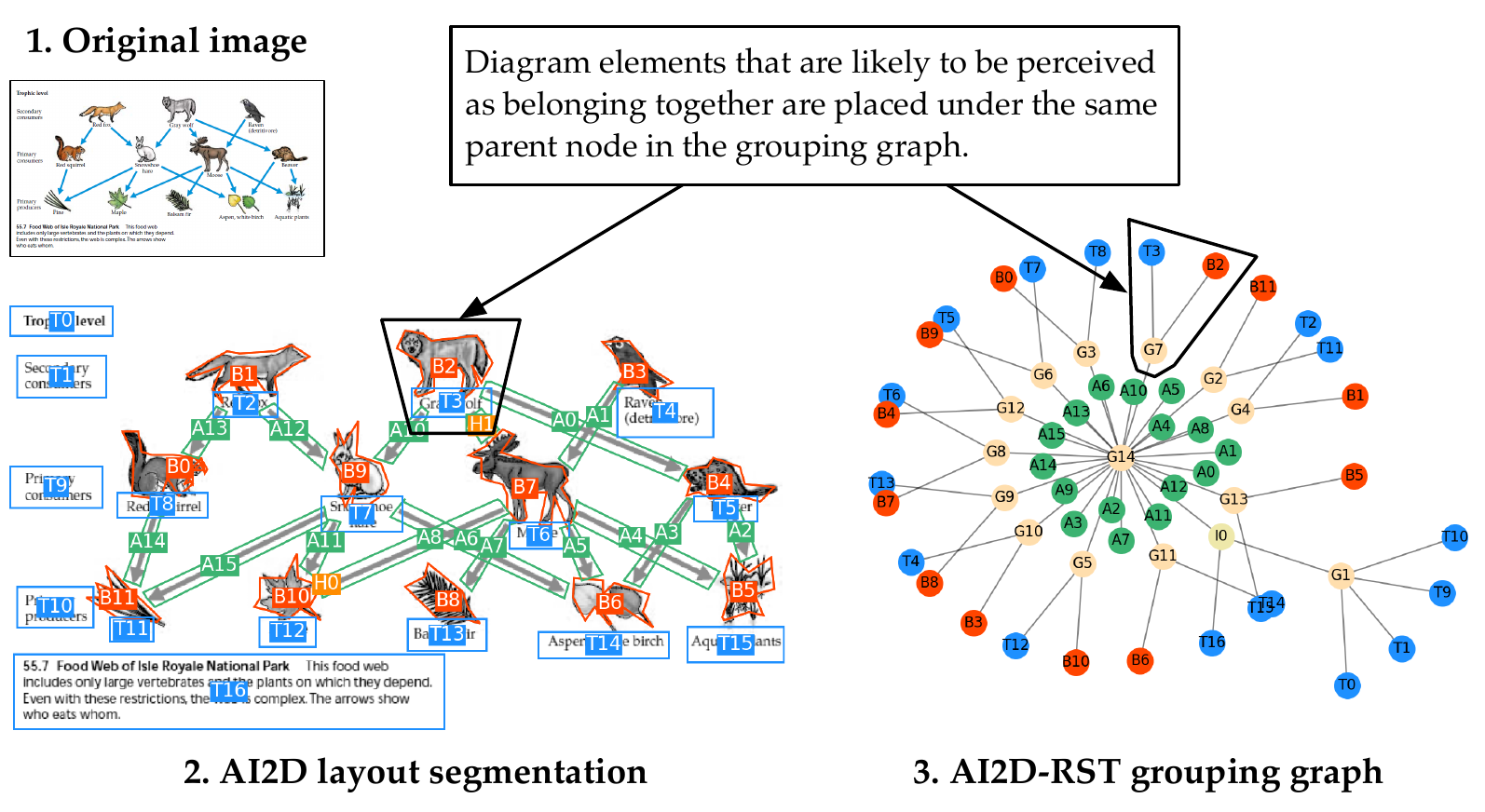}
    \caption{(1) A thumbnail of the original diagram image scraped from the web, (2) its crowd-sourced layout segmentation (converted into greyscale to bring out the annotation) and (3) the AI2D-RST grouping graph for diagram \#274. The grouping graph organises diagram elements that are likely to be perceived as belonging together into groups. These groups are added to the grouping graph as parent nodes for the diagram elements that belong together. For an example, see the illustration of a wolf (\emph{B2}) and the text `Gray wolf' (\emph{T3}) in the layout segmentation and their corresponding nodes in the AI2D-RST grouping graph). Both \emph{B2} and \emph{T3} are children of the grouping node \emph{G7}, which can be used to refer to both diagram elements in connectivity and discourse structure annotation.}
    \label{fig:grouping}
\end{figure}

The grouping layer describes which diagram elements form visual groups, that is, which elements are likely to be perceived as belonging together. The principles behind grouping correspond loosely to Gestalt principles of perception, which often act as guiding principles for designing diagrams and other visualisations \citep[179]{ware2012}. The grouping annotation is represented using an undirected, acyclic tree graph, such as the one shown on the right-hand side in Figure \ref{fig:grouping}.

In Figure \ref{fig:grouping}, the root node of the graph is the image constant \emph{I0}, which stands for the entire diagram. In contrast to AI2D, the AI2D-RST grouping layer includes nodes for only three types of diagram elements, namely blobs, text and arrows, but introduces another node type: groups. Diagram elements that form a visual unit in the layout are placed under the same parent node in the grouping graph. These nodes have the prefix G in their identifier, which stands for a group.

Conversely, besides grouping elements together, the grouping graph also represents which elements are considered independent, or in other words, do not belong to any visual groups. In Figure \ref{fig:grouping}, such independent units include the arrows \emph{A0--15} that set up the network of connections between the groups of illustrations and their labels \emph{G2--13}. These connections are described in the connectivity layer in order to avoid making arbitrary decisions about whether arrows should be grouped with their sources or targets (see Section \ref{sec:connectivity}).

To summarize, the grouping graph provides a foundation for the subsequent annotation layers, namely macro-grouping, connectivity and discourse structure \emph{by providing the necessary units of analysis}. In practice, the grouping graph allows diagram elements that form visual groups to be picked up for description in these layers by referring to the identifiers of their grouping nodes.

\subsubsection{Macro-grouping}
\label{sec:macrogrouping}

Macro-grouping captures the generic principles that govern diagram structure above the level of visual groups identified in the grouping layer, in order to describe why such visual groupings of expressive resources exist in the first place. To draw on an example, the grouping graph shown in Figure \ref{fig:grouping} consists of the groups \emph{G2--G13}, which combine an illustration and a written label, and the arrows \emph{A0--15}. Both groups and arrows form a single visual group, \emph{G14}, which may be appropriately characterised as a network. We term such constellations of visual groups \emph{macro-groups}, because they combine multiple visual groups into a larger structure.

Due to their close relation to the grouping layer, the annotation for macro-grouping layer is incorporated into the grouping graph. If the diagram consists of a single macro-group, macro-grouping information is assigned to the root node of the grouping graph (the image constant \emph{I0}), but if the diagram features multiple macro-groups, this information may be assigned to grouping nodes as well. Figure \ref{fig:grouping} exemplifies a diagram with multiple macro-groups. The food web under grouping node \emph{G14} is assigned the macro-group \emph{network}, whereas the categories on the left under the grouping node \emph{G1} form a \emph{vertical} organisation, whose function is to provide labels for visual groups in the network.

\begin{figure}[h!]
    \centering
    \includegraphics[width=0.7\textwidth]{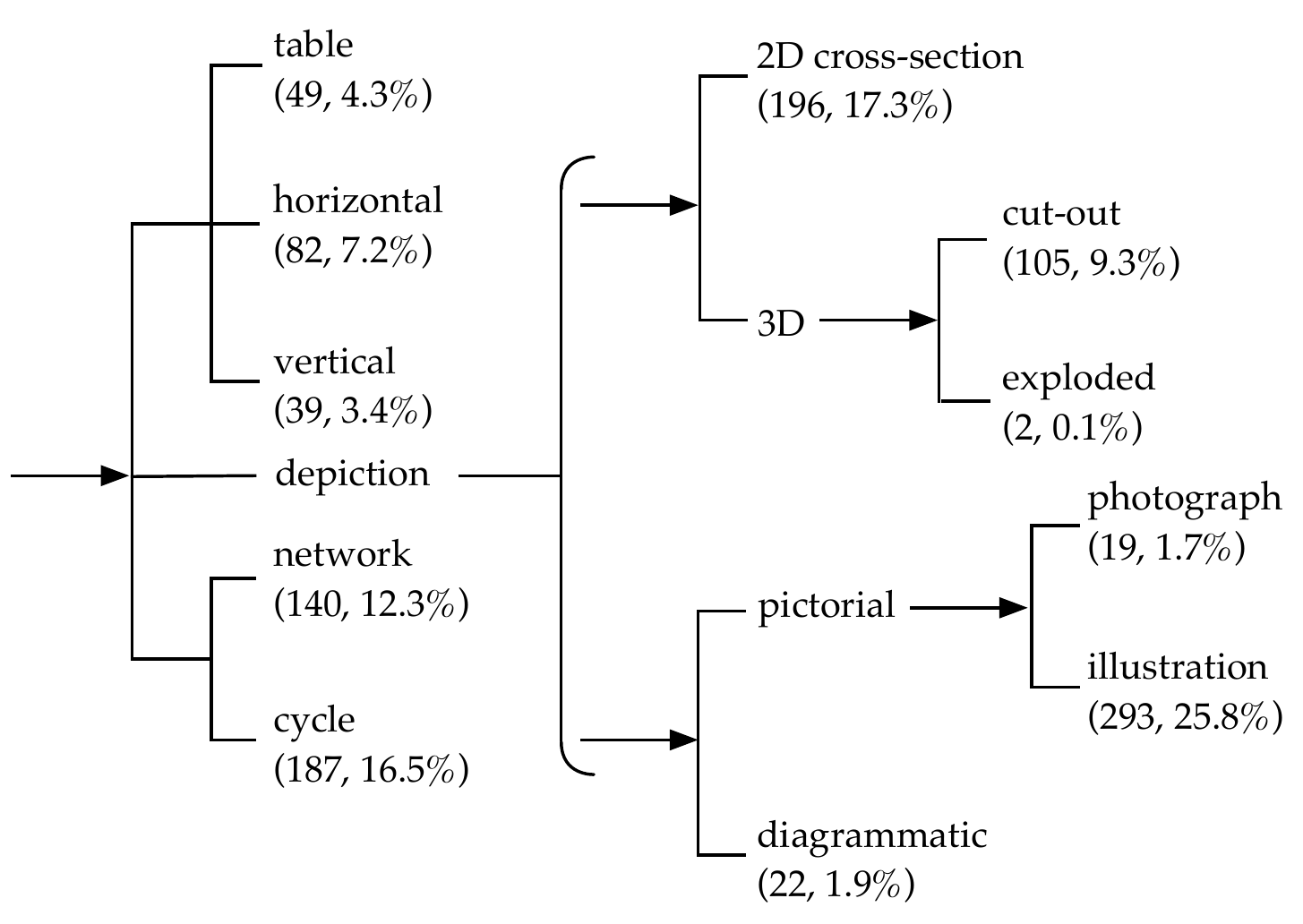}
    \caption{A typology in the form of a system network, which represents choices among macro-groups in the AI2D dataset. An arrow indicates an entry point, whereas a curly bracket indicates an alternative. To exemplify, choosing depiction requires making another choice between 2D/3D view or pictorial-diagrammatic representation. The numbers in parentheses give the raw count for each macro-group and their percentage of the AI2D-RST corpus ($N = 1134$).}
    \label{fig:macrogroups}
\end{figure}

Figure \ref{fig:macrogroups} shows a typology of macro-groups developed on the basis of our initial analysis of diagram types in the AI2D-RST corpus. As such, the scope of the typology is not intended to cover the space of possibilities within the entire diagrammatic mode, but is limited to the diagrams in AI2D-RST. In addition to describing the larger organisations of visual groups, macro-groups are intended to provide a set of structural categories that correspond to different \emph{diagram types}, in contrast to the semantic categories in AI2D, which are based on the subject matter of the diagram. In this way, macro-groups can also be used as target labels for training classifiers.

\subsubsection{Connectivity}
\label{sec:connectivity}

The connectivity layer describes connections between diagram elements and their groups, which are signalled visually using arrows, lines and other diagrammatic elements capable of expressing connectivity \citep{tverskyetal2000}. In AI2D-RST, the connectivity annotation covers visually explicit connections between diagram elements only, that is, the arrows and lines must have a clear source and a target, in order to allow the connections to be represented using graphs \citep[cf.][3554]{alikhanistone2018}. The AI2D-RST annotation schema defines three types of connections based on directionality: undirected, directed and bidirectional.

\begin{figure}[h!]
    \centering
    \includegraphics[width=1\textwidth]{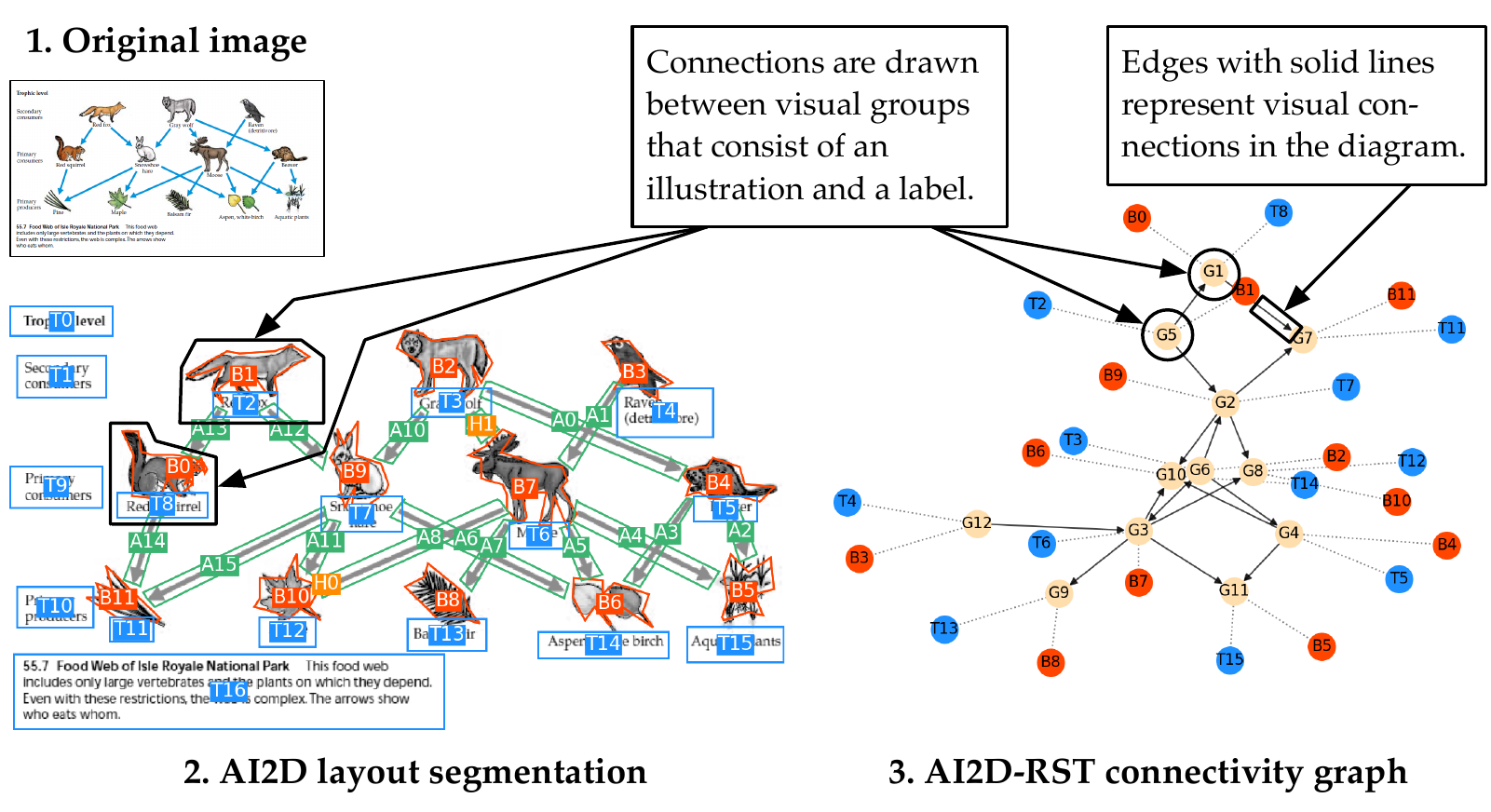}
    \caption{(1) A thumbnail of the original diagram image scraped from the web, (2) its crowd-sourced layout segmentation (converted into greyscale to bring out the annotation) and (3) the AI2D-RST connectivity graph for diagram \#274. In the connectivity graph, the edges with solid lines correspond to arrows in the layout segmentation, whereas edges with dashed lines represent edges in the \emph{grouping} graph, which show the diagram elements that form the visual group.}
    \label{fig:connectivity}
\end{figure}

The connectivity annotation is represented using a cyclic mixed graph, which means that the graph may feature both undirected and directed edges. Figure \ref{fig:connectivity} exemplifies a connectivity graph, whose visualization has been enhanced with edges from the grouping graph (for the original grouping graph, see Figure \ref{fig:grouping}), because the connections in Figure \ref{fig:connectivity} are likely to be perceived to hold between \emph{visual groups} of elements, rather than individual elements, such as labels or illustrations. Annotating connectivity according to visually explicit connections between individual elements, which originate and terminate in both labels and illustrations, as exemplified by the directed connection between the text block \emph{T3} (`Gray wolf') and the illustration of a hare in \emph{B9}, results in an incomplete representation of connectivity. This shows why visual groups are needed as basic units of analysis for a graph-based representation of connectivity, which also illustrates how the grouping layer supports other annotation layers by providing the necessary units of analysis.

\subsubsection{Discourse structure}
\label{sec:discourse}

Whereas the grouping and connectivity layers seek to capture the diagram structure that is \emph{explicitly} available for visual inspection, the discourse structure layer attempts to describe the \emph{implicit} discourse relations that hold between diagram elements and their groups, which viewers may recover from the diagram structure. As such, the discourse structure layer provides the crucial link between multimodal structure and communicative intentions.

\begin{figure}[h!]
    \centering
    \includegraphics[width=1\textwidth]{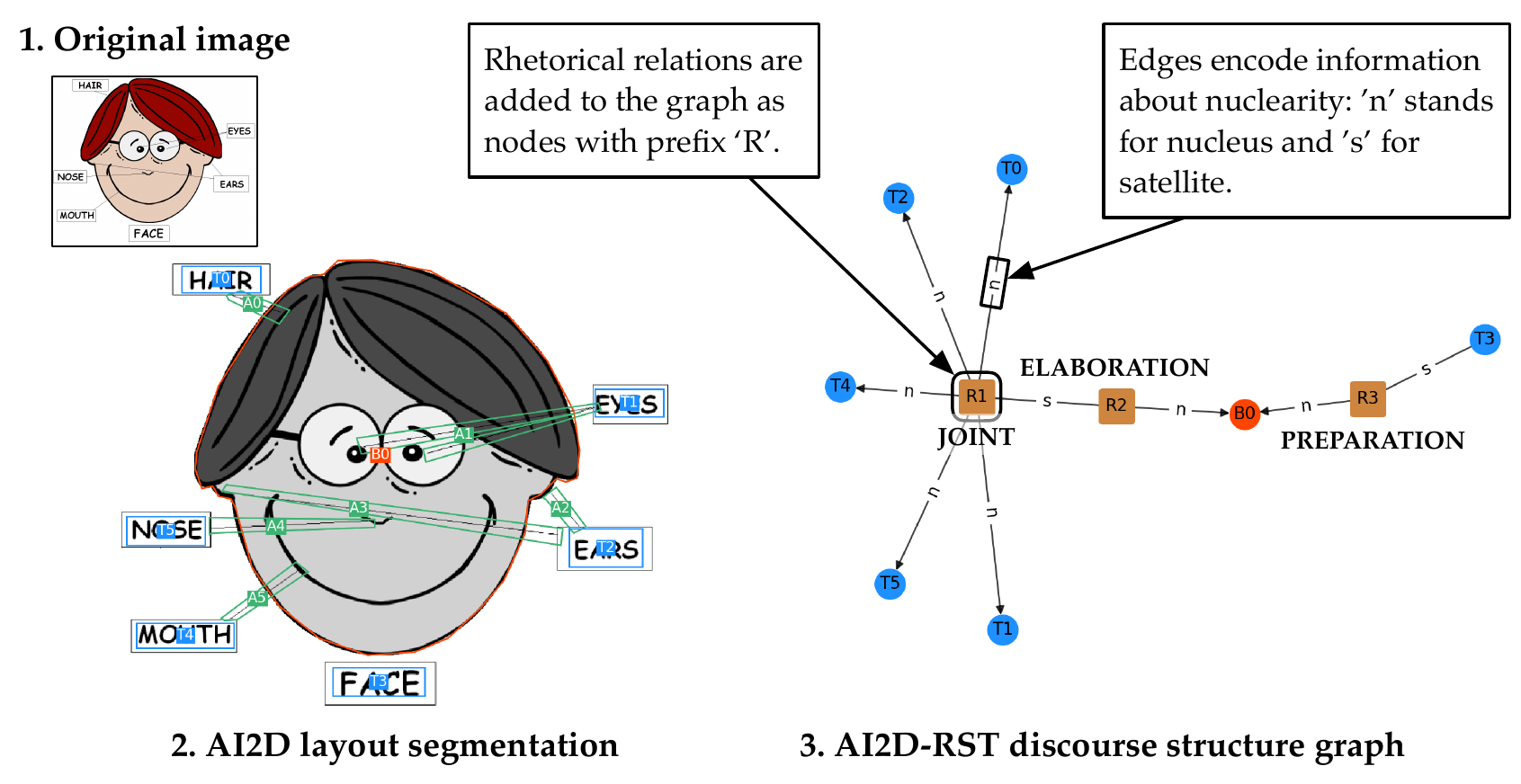}
    \caption{(1) A thumbnail of the original diagram image scraped from the web, (2) its crowd-sourced layout segmentation (converted into greyscale to bring out the annotation) and (3) the AI2D-RST discourse structure graph for diagram \#0. The multinuclear \textsc{joint} relation \emph{R1} joins together the labels \emph{T0--2} and \emph{T4--5}, which serve a similar communicative purpose in the diagram, that is, pick out parts of the illustration \emph{B0} for description. Part-whole relations are described using the \textsc{elaboration} relation \emph{R2}, in which the \textsc{joint} relation \emph{R1} acts as a satellite and the illustration \emph{B0} as the nucleus. Another relation on the highest level of the hierarchy is drawn between the illustration \emph{B0} and the text \emph{T3} (`FACE') that describes the entire diagram, which is annotated as \textsc{preparation} (\emph{R3}). The edge labels `n' and `s' stand for nucleus and satellite, respectively.}
    \label{fig:rst_0}
\end{figure}

For describing the discourse structure of diagrams, AI2D-RST uses Rhetorical Structure Theory \citep[RST; see e.g.][]{mannthompson1988, taboadamann2006a}, a theory of textual organisation and coherence which has been previously extended to diagrams in natural language generation \citep{andrerist1995, batemanetal2001, batemanhenschel2007} and for describing discourse relations in research on multimodal documents and other artefacts \citep{bateman2008, thomas2009a, taboadahabel2013, hiippala2015a}. This extension of RST, which may be described as \emph{multimodal RST}, provides the foundation for discourse structure annotation in AI2D-RST, as exemplified in Figure \ref{fig:rst_0}.

\begin{figure}[h!]
    \centering
    \includegraphics[width=1\textwidth]{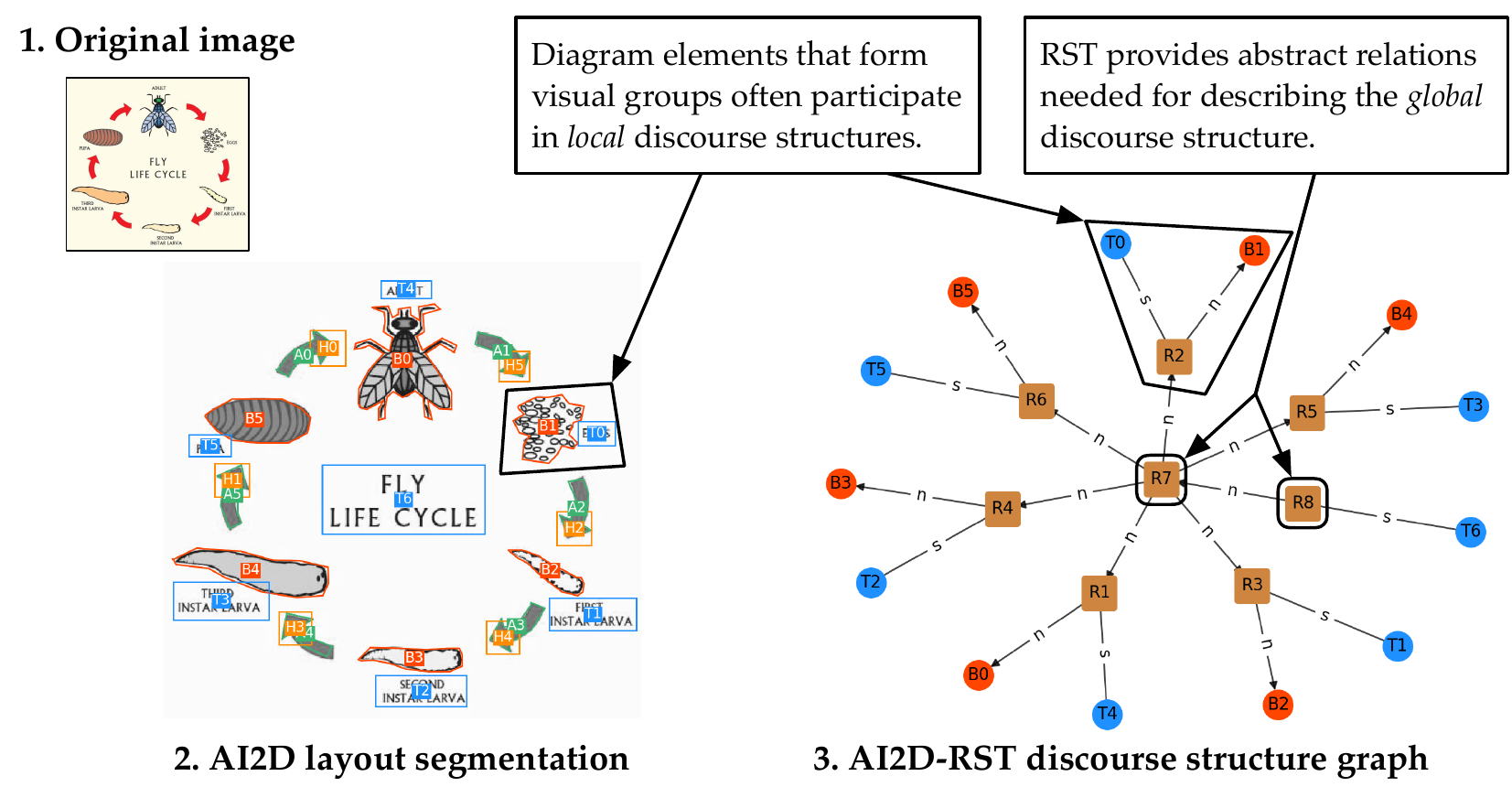}
    \caption{(1) A thumbnail of the original diagram image scraped from the web, (2) its crowd-sourced layout segmentation (converted into greyscale to bring out the annotation) and (3) the AI2D-RST discourse structure graph for diagram \#2185. The diagram features three distinct types of rhetorical relations. Whereas the \textsc{identification} relations (\emph{R1--6}) are mainly local in the sense that the participating diagram elements form visual groups, the \textsc{cyclic sequence} (\emph{R7}) and \textsc{preparation} (\emph{R8}) describe the global discourse organisation of the diagram, or how larger formations of discourse units relate to each other.}
    \label{fig:rst_2185}
\end{figure}

Both `classical' and multimodal RST provide a set of discourse relations with criteria for their application \citep{mannthompson1988,bateman2008}. For annotating discourse relations in the AI2D-RST corpus, we used the relation definitions presented in \citet[221--223]{hiippala2015a} which cover the classical RST relations from \citet{mannthompson1988} with the multimodal extension proposed in \citet{bateman2008}. We also introduced an additional relation, \textsc{cyclic sequence}, which is used to describe repeating sequences (see the example in Figure \ref{fig:rst_2185}).

We drew on these relation definitions to describe how elementary discourse units -- which in AI2D-RST correspond to diagram elements or their groups -- relate to each other. Depending on the relation, one discourse unit may be considered nuclear, or more important, whereas other units act as satellites that play a secondary role. RST terms such relations \emph{asymmetric}. \emph{Symmetric} relations, in turn, may have multiple nuclei, indicating equal status among discourse units. The example in Figure \ref{fig:rst_0} exemplifies both symmetric (\emph{R2}, \emph{R3}) and asymmetric (\emph{R1}) relations and illustrates how RST relations are represented in the discourse structure graph. Relations are added to the graph as nodes whose identifier has the prefix R, whereas the edges between these nodes carry information on nuclearity, that is, whether the participating diagram elements act as nuclei or satellites.

Figure \ref{fig:rst_2185} shows a more complex example, which illustrates the benefit of adopting RST for describing the discourse structure of diagrams. As pointed out above in Section \ref{sec:ai2d}, the relation definitions in the AI2D annotation schema are largely constrained to local relations between adjacent elements. RST, in turn, provides abstract relations that can handle the description of global discourse organisation as well, or how larger constellations of diagram elements relate to each other.

RST analyses are commonly represented using recursive tree diagrams, although this is not a requirement set by the theory \citep[435]{taboadamann2006a}. \citet{wolfgibson2005} have argued that tree structures are too constrained for an accurate representation of discourse structure, because a single discourse unit may be picked up as a part of multiple discourse relations. They propose using graphs as an alternative data structure, which would allow discourse units to participate in multiple relations and abolish the hierarchical tree structure.

\begin{figure}[h!]
    \centering
    \includegraphics[width=1\textwidth]{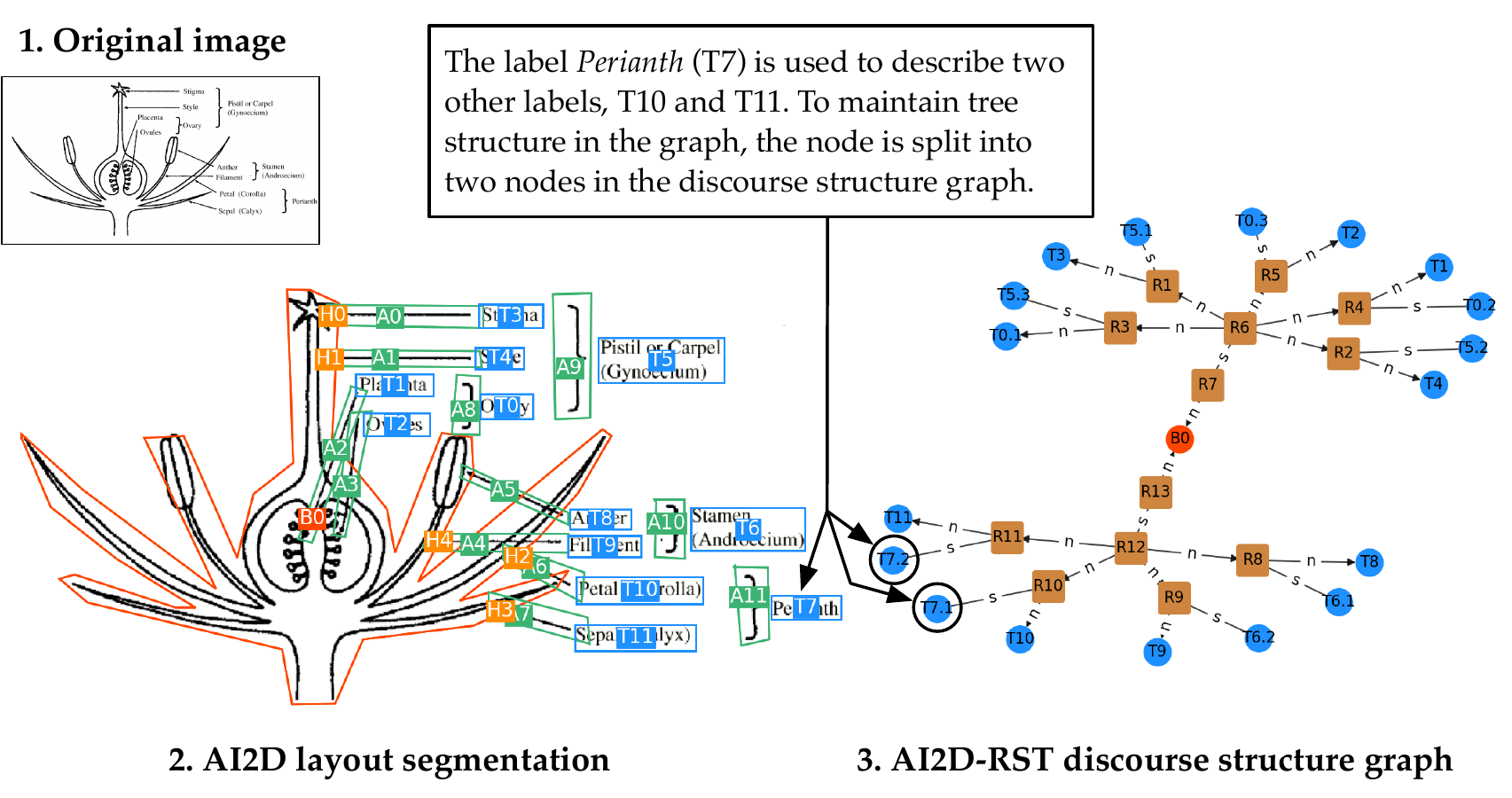}
    \caption{(1) A thumbnail of the original diagram image scraped from the web, (2) its crowd-sourced layout segmentation and (3) the AI2D-RST discourse structure graph for diagram \#3194. The diagram features three distinct types of rhetorical relations: \textsc{identification} (\emph{R1--5}, \emph{R8--11}), \textsc{elaboration} (\emph{R7}, \emph{R13}) and \textsc{joint} (\emph{R6}, \emph{R12}). To preserve the tree structure of the graph, several diagram elements are represented by multiple nodes in the discourse structure graph, as these elements participate in multiple rhetorical relations. To exemplify, the label \emph{Perianth} (\emph{T7}) describes two other labels, \emph{Petal (Corolla)} (\emph{T10}) and \emph{Sepal (Calyx)} (\emph{T11}). We describe this relation as \textsc{identification}, as the label \emph{T7} identifies that the labels \emph{T10} and \emph{T11} collectively form a part named \emph{Perianth}. In the discourse structure graph, the \textsc{identification} relations (\emph{R10} and \emph{R11}) both feature a copy of \emph{T7} as a satellite to preserve the tree structure.}
    \label{fig:rst_3194}
\end{figure}

The discourse structure layer, however, preserves the hierarchical structure and uses a directed acyclic tree graph to represent RST analyses. This decision is motivated by the use of layout space in diagrams, which is regularly used to set up discourse relations between diagram elements \citep{waller2012,watanabenagao1998}. The inherently spatial organisation of diagrams makes constraining the application of discourse relations difficult, particularly in terms of spatial adjacency, that is, limiting relations to elements that are positioned close to each other \citep[cf.][158]{bateman2008}. We consider preserving the tree structure to impose additional control on the application of RST relations.

We do, however, acknowledge that like multimodal documents, diagrams can `re-use' discourse units in different rhetorical relations \citep[159]{bateman2008}. To account for diagram elements that take on the role of satellites or nuclei in multiple rhetorical relations, we split the diagram elements to preserve the hierarchical structure, as shown in Figure \ref{fig:rst_3194}. This involves creating copies of a node in the graph, which are identified using a decimal in the node name, such as \emph{T7.1} or \emph{T7.2}. Each copy of the node may be then picked up in the RST analysis while preserving the tree structure. Because the original identifiers are preserved as attributes of the split nodes in the discourse structure graph, the acyclic tree graphs can be easily converted into cyclic graphs favoured by \citet{wolfgibson2005}, if necessary.

\subsection{Annotators and training}

The AI2D-RST diagrams were annotated by five students pursuing BA or MA degrees in English, who received approximately 10 hours of initial training in the form of introductory sessions covering each annotation layer. They also received detailed feedback on their initial work and could pose questions about the application of the annotation schema using an online tool for team collaboration. The annotators were also supported by a document that provided guidelines and preferred solutions to common annotation problems, which is available in the repository associated with this article. We return to discuss the impact that the collaborative annotation process may have had on the reproducibility of the annotation framework at the end of Section \ref{sec:reliability}.

\subsection{The annotation tool}

We developed an in-house tool to annotate the diagrams. The tool provides a command line interface for building graphs, which are initially populated by nodes from the original AI2D layout segmentation. The tool is written in Python 3.6 and makes extensive use of the \emph{matplotlib} \citep{hunter2007}, \emph{NetworkX} \citep{hagbergetal2008}, \emph{OpenCV} \citep{bradski2013} and \emph{pandas} \citep{mckinney2010} libraries. The tool and its source code are available with an open license at \url{https://doi.org/10.5281/zenodo.3384751}.

\subsection{Acquiring the corpus}

The AI2D-RST corpus is available for download as JSON files in the Language Bank of Finland: \url{http://urn.fi/urn:nbn:fi:lb-2019120407}. Python functions for loading and processing the corpus are provided separately at \url{https://doi.org/10.5281/zenodo.3384751}.

\section{Measuring the reliability of the annotation}
\label{sec:reliability}

We measured inter-annotator agreement when 355 diagrams had been annotated. At this stage, the annotators were assumed to have familiarised themselves with the annotation schema. Because the data was annotated by five annotators, we used Fleiss' kappa ($\kappa$) as implemented in the \emph{statsmodels} Python library \citep{seaboldperktold2010} as the metric for measuring inter-annotator agreement. We report both the original $\kappa$ statistic, as proposed by \citet{fleiss1971}, which is calculated using marginal probabilities for each category, and the free-marginal $\kappa$ proposed by Randolph \citeyearpar{randolph2005}, which assumes a uniform distribution over all categories. We refer to Fleiss' original definition as marginal $\kappa$ and Randolph's alternative as uniform $\kappa$. In addition, we used the \emph{irr} library \citep{gameretal2019} for the R programming language \citep{r2019} to calculate class-wise marginal $\kappa$ scores for grouping, macro-grouping, connectivity and discourse structure annotations. The results are reported in Sections \ref{sec:rel-group}, \ref{sec:rel-macro}, \ref{sec:rel-conn} and \ref{sec:rel-rst}. Finally, in Section \ref{sec:comp} we model annotator reliability using MACE \citep{hovyetal2013}. The raw annotations are provided as CSV files at \url{https://doi.org/10.5281/zenodo.3384751}.

\subsection{Grouping} 
\label{sec:rel-group}

\begin{table}[h!]
\centering
\caption{Class-wise marginal $\kappa$ scores for Gestalt principles and annotation guidelines} \vspace{2mm}
\label{tab:gestalt-groups}
\begin{tabular}{l p{7.5cm} r r r}
\hline\noalign{\smallskip}
Category & Description & $\kappa$ & z-score & p-value \\
\noalign{\smallskip}\hline\noalign{\smallskip}
Guideline & The AI2D-RST guidelines state that the elements are grouped together. & 0.929 & 47.008 & $<0.001$ \\
Proximity & The diagram elements are placed close to each other in the layout space. & 0.851 & 43.046 & $<0.001$ \\
Closure & The element encloses the other element. & 0.776 & 39.243 & $<0.001$\\
Similarity & The elements are similar in terms of their visual appearance. & 0.622 & 31.453 & $<0.001$ \\
No-group & The elements do not form a valid group according to the AI2D-RST schema. & 0.410 & 20.766 & $<0.001$ \\
Continuity & The elements form a continuous unit. & 0.210 & 10.623 & $<0.001$ \\
Connectedness & The elements are connected to each other. & -0.003 & -0.159 & 0.874 \\
Symmetry & The elements form a symmetrical shape. & -0.002 & -0.079 & 0.937\\
\noalign{\smallskip}\hline\noalign{\smallskip}
\end{tabular}
\end{table}

To evaluate the reliability of grouping layer annotation introduced in Section \ref{sec:grouping}, we sampled the 355 completed diagrams without replacement for 10\% of visual groups composed of diagram elements only, excluding groups whose child nodes included other grouping nodes. This amounted to 256 groups, whose elements were highlighted in the AI2D layout segmentation and presented to the annotators. The annotators were then asked whether the elements form a visual group, as defined in the AI2D-RST annotation schema. If the annotators considered the grouping valid, a follow-up question requested the annotators to name Gestalt principle or annotation guideline that justified their choice. If multiple principles or guidelines were applicable, the annotators were asked to choose the most prominent one. For inter-annotator agreement between five annotators and 256 groups, the marginal $\kappa$ was 0.836, while the uniform $\kappa$ was 0.894.

Table \ref{tab:gestalt-groups} shows class-wise agreement for Gestalt principles and annotation guidelines, which are sorted in descending order based on their marginal $\kappa$ values. The results suggest that the annotation guide supported the consistent description of the data. Most cases in the guideline category consisted of label--line combinations, such as those shown in Figure \ref{fig:rst_0}. In principle, such combinations could be grouped together based on several Gestalt principles, such as proximity, continuity and connectedness, but explicating annotation patterns for common diagrammatic structures such as labels and their connecting lines seems to make the decisions less arbitrary. In addition, common spatial- and attribute-based relations that build on Gestalt principles such as proximity, closure and similarity \citep[30]{engelhardt2002}, are annotated consistently in the AI2D-RST corpus.

\subsection{Macro-grouping} 
\label{sec:rel-macro}

For measuring inter-annotator agreement on the macro-groups introduced in Section \ref{sec:macrogrouping}, we sampled the 355 completed diagrams without replacement for 33\% of macro-groups, which amounted to 119 macro-groups. The annotators were presented with the AI2D layout segmentation and the AI2D-RST grouping graph, which highlighted the node that had been assigned with macro-grouping information. The annotators were then asked which macro-group they would assign to the node in question. For inter-annotator agreement on macro-groups, the marginal $\kappa$ was 0.784 and the uniform $\kappa$ was 0.800.

\begin{table}[h!]
\centering
\caption{Class-wise marginal $\kappa$ scores for macro-groups} \vspace{2mm}
\label{tab:rel-macro-groups}
\begin{tabular}{l r r r r}
\hline\noalign{\smallskip}
Macro-group & $\kappa$ & z-score & p-value & Frequency in the entire corpus \\
\noalign{\smallskip}\hline\noalign{\smallskip}
Network & 0.884 & 30.480 & $<0.001$ & 0.123\\
Cycle & 0.876 & 30.204 & $<0.001$ & 0.165 \\
Cut-out & 0.849 & 29.271 & $<0.001$ & 0.093 \\
Slice & 0.754 &25.996 & $<0.001$ & 0.173 \\
Horizontal & 0.726 & 25.031 & $<0.001$ & 0.072 \\
Diagrammatic & 0.718 & 24.785 & $<0.001$ & 0.019 \\
Illustration & 0.709 & 24.458 & $<0.001$ & 0.258 \\
Vertical & 0.702 & 24.228 & $<0.001$ & 0.034 \\
Table & 0.247 & 8.537 & $<0.001$ & 0.043 \\
Photograph & 0.162 & 5.604 & $<0.001$ & 0.017 \\
\noalign{\smallskip}\hline\noalign{\smallskip}
\end{tabular}
\end{table}

Table \ref{tab:rel-macro-groups} gives class-wise marginal $\kappa$ values for macro-groups in descending order. Agreement is particularly high for visually distinctive macro-groups such as networks, cycles and cut-outs, which occur frequently in the AI2D-RST corpus (see also Figure \ref{fig:macrogroups}). The values are considerably lower for less common macro-groups such as tables and photographs. Photographs, in particular, are rarely preferred as the main visual mode of expression in the AI2D-RST corpus, as diagrams in the corpus favour illustrations, cut-outs and cross-sections for visual expression. For these prominent macro-groups, agreement remains substantial.

\subsection{Connectivity} 
\label{sec:rel-conn}

For connectivity annotation (see Section \ref{sec:connectivity}), we sampled the 355 completed diagrams without replacement for 10\% connections holding between diagram elements or their groups, which resulted in 239 connections. The source and target of each connection were highlighted in the AI2D layout segmentation and presented to the annotators, who were then asked to place the connection into one of four categories: directed, undirected, bidirectional or no connection. Measuring inter-annotator agreement returned a marginal $\kappa$ of 0.878 and uniform $\kappa$ of 0.916. Table \ref{tab:rel-connectivity} gives class-wise marginal $\kappa$ values for each connection type. Apart from no connection, agreement is high across all types of connectivity, as might be expected with a low number of categories, which are also visually distinctive and whose structural features are relatively easy to formalize \citep[see][3554]{alikhanistone2018}.

\begin{table}[h!]
\centering 
\caption{Class-wise marginal $\kappa$ scores for connectivity} \vspace{2mm}
\label{tab:rel-connectivity}
\begin{tabular}{l l l l r}
\hline\noalign{\smallskip}
Connection & $\kappa$ & z-score & p-value & Frequency in the entire corpus \\
\noalign{\smallskip}\hline\noalign{\smallskip}
Directed & 0.910 & 44.512 & $<0.001$ & 0.511 \\
Bidirectional & 0.908 & 44.402 & $<0.001$ & 0.004 \\
Undirected & 0.900 & 44.003 & $<0.001$ & 0.485 \\
No connection & 0.192 & 9.392& $<0.001$ & N/A \\
\noalign{\smallskip}\hline\noalign{\smallskip}
\end{tabular}
\end{table}

\subsection{Discourse structure} 
\label{sec:rel-rst}

For evaluating inter-annotator agreement on the discourse structure layer introduced in Section \ref{sec:discourse}, we sampled the 355 completed diagrams without replacement for 10\% of the relations, amounting to a total of 227 RST relations. The AI2D layout segmentation and the AI2D-RST discourse structure graph were presented side-by-side to the annotators, highlighting the RST relation node to be annotated in the discourse structure graph. Measuring overall agreement on the RST relations returned a marginal $\kappa$ of 0.733 and a uniform $\kappa$ of 0.783.

\begin{table}[h!]
\centering
\caption{Class-wise marginal $\kappa$ scores for discourse relations}  \vspace{2mm}
\label{tab:rel-discourse}
\begin{tabular}{l r r r r}
\hline\noalign{\smallskip}
Discourse relation & $\kappa$ & z-score & p-value & Frequency in the entire corpus \\
\noalign{\smallskip}\hline\noalign{\smallskip}
\textsc{cyclic sequence} & 0.924 & 44.029 & $<0.001$ & 0.033 \\
\textsc{preparation} & 0.870 & 41.471 & $<0.001$ & 0.054 \\
\textsc{property-ascription} & 0.870 & 41.468 & $<0.001$ & 0.070 \\
\textsc{joint} & 0.827 & 39.419 & $<0.001$ & 0.109 \\
\textsc{identification} & 0.798 & 37.998 & $<0.001$ & 0.439 \\
\textsc{connected} & 0.766 & 36.492 & $<0.001$ & 0.030 \\
\textsc{sequence} & 0.689 & 32.844 & $<0.001$ & 0.015 \\
\textsc{elaboration} & 0.620 & 29.540 & $<0.001$ & 0.134 \\
\textsc{circumstance} & 0.449 & 21.388 & $<0.001$ & 0.029 \\
\textsc{contrast} & 0.308 & 14.656 & $<0.001$ & 0.024 \\
\textsc{class-ascription} & 0.266 & 12.680& $<0.001$ & 0.028 \\
\textsc{conjunction} & 0.249 & 11.848 & $<0.001$ & $0.003$ \\
\textsc{disjunction} & 0.249 & 11.848 & $<0.001$ & 0.003 \\
\textsc{list} & 0.182 & 8.659 & $<0.001$ & 0.007 \\
\textsc{nonvolitional cause} & 0.138 & 6.553 & $<0.001$ & 0.004 \\
\textsc{nonvolitional result} & 0.078 & 3.738 & $<0.001$ & 0.006 \\
\textsc{means} & 0.066 & 3.129 & 0.002 & 0.003 \\
\textsc{condition} & -0.001 & -0.042 & 0.966 & 0.001 \\
\textsc{purpose} & -0.001 & -0.042 & 0.966 & N/A \\
\textsc{restatement} & -0.003 & -0.126 & 0.900 & 0.004 \\
\noalign{\smallskip}\hline\noalign{\smallskip}
\end{tabular}
\end{table}

Table \ref{tab:rel-discourse} provides class-wise marginal $\kappa$ scores for RST relations that the annotators used during the inter-annotator agreement experiment in a descending order. The results show that annotators consistently agree on common RST relations such as \textsc{cyclic sequence}, which is used to annotate recurring cycles formed by diagram elements, and \textsc{preparation}, which is used to describe the relationship between a title and an entire diagram. These RST relations are associated with visually distinctive \textsc{macro-groups} (cycles) and relatively fixed diagram elements (titles), which is likely to increase agreement. The same applies to frequently occurring relations defined between a label and an object or its part, such as \textsc{property-ascription}, \textsc{identification} and \textsc{elaboration}, whose specific use cases were defined in the annotation guide. In short, the development of an annotation guide seemed to support the consistent annotation of RST relations. Compared to previous studies of inter-annotator agreement using multimodal RST \citep[e.g.][]{taboadahabel2013}, the $\kappa$ scores for the AI2D-RST discourse structure layer are promising, as relations with a $\kappa > 0.62$ cover 88.4\% of RST relations in the corpus.

\begin{figure}[h!]
    \centering
    \includegraphics[width=1\textwidth]{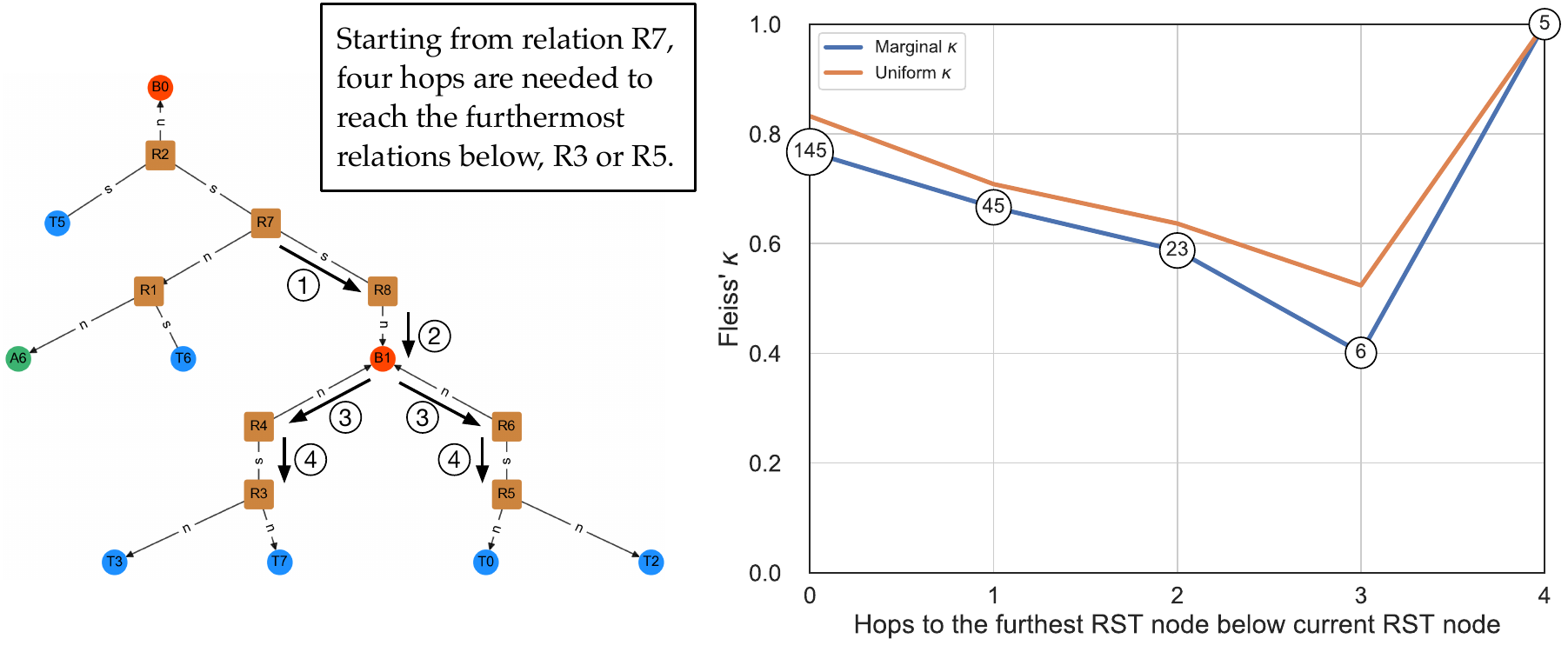}
    \caption{Fleiss' marginal and uniform $\kappa$ for RST relations at different depths of the RST tree. We measured the position of RST nodes in the tree by calculating the number of hops needed to reach the furthermost RST node in the subtree below, as illustrated on the left-hand side. On the right-hand side, the balloons give the number of samples observed for each hop. The X-axis gives the number of hops: a value of zero indicates that the RST relation is close to the edge of the tree.}
    \label{fig:rst_reliability}
\end{figure}

Figure \ref{fig:rst_reliability} provides an alternative view to the reliability of the discourse structure annotation by measuring inter-annotator agreement at different depths of the RST tree graph. Not surprisingly, agreement is highest at the leaves of the tree graph (hop 0) with a marginal $k$ of 0.767 and a uniform $k$ of 0.832. These consistently annotated relations mainly cover local discourse structures illustrated in Figure \ref{fig:rst_2185}, as exemplified by \textsc{identification} ($N = 81$), \textsc{joint} ($N = 21$) and \textsc{property-ascription} ($N = 21$). As the $\kappa$ values for hops 1--3 show, agreement decreases for relations that are positioned up the tree, which represent the more abstract relations that hold between larger discourse units. Surprisingly, annotators consistently agree on how the relations closest to the root (hop 4) should be annotated. It should be noted, however, that sample sizes are very small for hops 3 and 4 and should therefore be treated with caution.

\subsection{Modelling annotator reliability} \label{sec:comp}

In addition to measuring inter-annotator agreement, we estimated annotator reliability using MACE \citep{hovyetal2013}. MACE, which stands for Multi-Annotator Competence Estimation, models the annotation process by treating the labels as latent variables and uses unsupervised learning to estimate the model parameters. The model seeks to predict whether the annotator is answering dutifully or choosing the answers at random. \citet[1124]{hovyetal2013} show that MACE reliability estimates correlate strongly with annotator proficiency. Table \ref{tab:mace} shows MACE reliability estimates using default settings, which suggests dutiful annotation with slightly varying competences between annotators. 

\begin{table}[h!]
\centering
\caption{MACE reliability estimates for annotators and specific tasks}  \vspace{2mm}
\label{tab:mace}
\begin{tabular}{l r r r r r}
\hline\noalign{\smallskip}
Task & Ann. 1 & Ann. 2 & Ann. 3 & Ann. 4 & Ann. 5 \\
\noalign{\smallskip}\hline\noalign{\smallskip}
Grouping & 0.9133 & 0.9378 & 0.9040 & 0.9601 & 0.9430 \\
Macro-grouping & 0.8851 & 0.8052 & 0.9351 & 0.8574 & 0.8954 \\
Connectivity & 0.9478 & 0.9382 & 0.9531 & 0.9364 & 0.9631 \\
Discourse structure & 0.8452 & 0.8698 & 0.8912 & 0.8021 & 0.9249 \\
\noalign{\smallskip}\hline\noalign{\smallskip}
\end{tabular}
\end{table}

\subsection{On the reliability and reproducibility of the AI2D-RST annotation schema}

Overall, the results for measuring inter-annotator agreement suggest that the AI2D-RST annotation is applied consistently to the diagrams. The results are particularly promising given that inter-annotator agreement was measured between five annotators. However, it is important to acknowledge that measuring inter-annotator agreement using metrics such as Fleiss' $\kappa$ often involve compromises. In the case of RST, for instance, measuring agreement over a single relation in a given context is very different from constructing entire RST trees and comparing them between annotators. To improve the evaluation of annotation reliability, future studies applying multimodal RST should follow up on recent developments in research on the automatic comparison of RST trees \citep[see e.g.][]{wanetal2019}. Alternatively, the approach illustrated in Figure \ref{fig:rst_reliability} could be used sample relations along the depth of the RST tree in a balanced manner, in order to ensure that agreement is evaluated for both local and global discourse structures.

In terms of the annotation schema, it should be noted that the expert annotators helped to develop the AI2D-RST annotation schema by discussing specific examples with each other, which were then documented in the annotation guide. This violates several principles of reproducibility set out for \emph{content analysis} in \citet{krippendorff2013}. However, as \citet[575]{artsteinpoesio2008} point out, content analysis treats the annotation process as an \emph{experiment} about whether some properties may be consistently detected in a text, whose success is determined by reproducibility of the annotation. In computational linguistics, annotation serves a different purpose, such as creating resources for training and evaluating algorithms, which differs from the goals of content analysis \citep{reidsmacarletta2007}.

\citet[240]{riezler2014}, however, also calls for attention to the consequences of violating the requirement of independence, that is, allowing the annotators to discuss annotation tasks. This is likely to generate implicit knowledge among the annotators, which increases agreement among annotators but hinders reproducibility. This kind of implicit knowledge gives rise to circularity in annotation, which has been acknowledged as a problem in multimodality research \citep{thomas2014}. Given the collaborative annotation procedure, it is likely that the AI2D-RST annotations exhibit a degree of circularity.

To evaluate and improve the reproducibility of the AI2D-RST framework, future work should employ naive annotators, who are assigned tasks that do not build on concepts introduced in the annotation framework \cite[see e.g.][]{asheghietal2016}. This kind of non-theoretical grounding \citep{riezler2014} could help to break circularity by evaluating, for instance, whether naive annotators perceive diagram elements to form visual groups (grouping) or whether arrows and lines are considered to signal connections between individual diagram elements or visual groups (connectivity). For discourse structure annotation, \cite{yungetal2019} introduce a multi-step procedure for sourcing descriptions of discourse relations from naive annotators. Adopting this approach in multimodal RST, however, would require additional efforts to accommodate the presence of multiple expressive resources.

\section{Exploring the AI2D-RST corpus}
\label{sec:description}

\begin{figure}[p]
    \centering
    \includegraphics[width=0.8\textwidth]{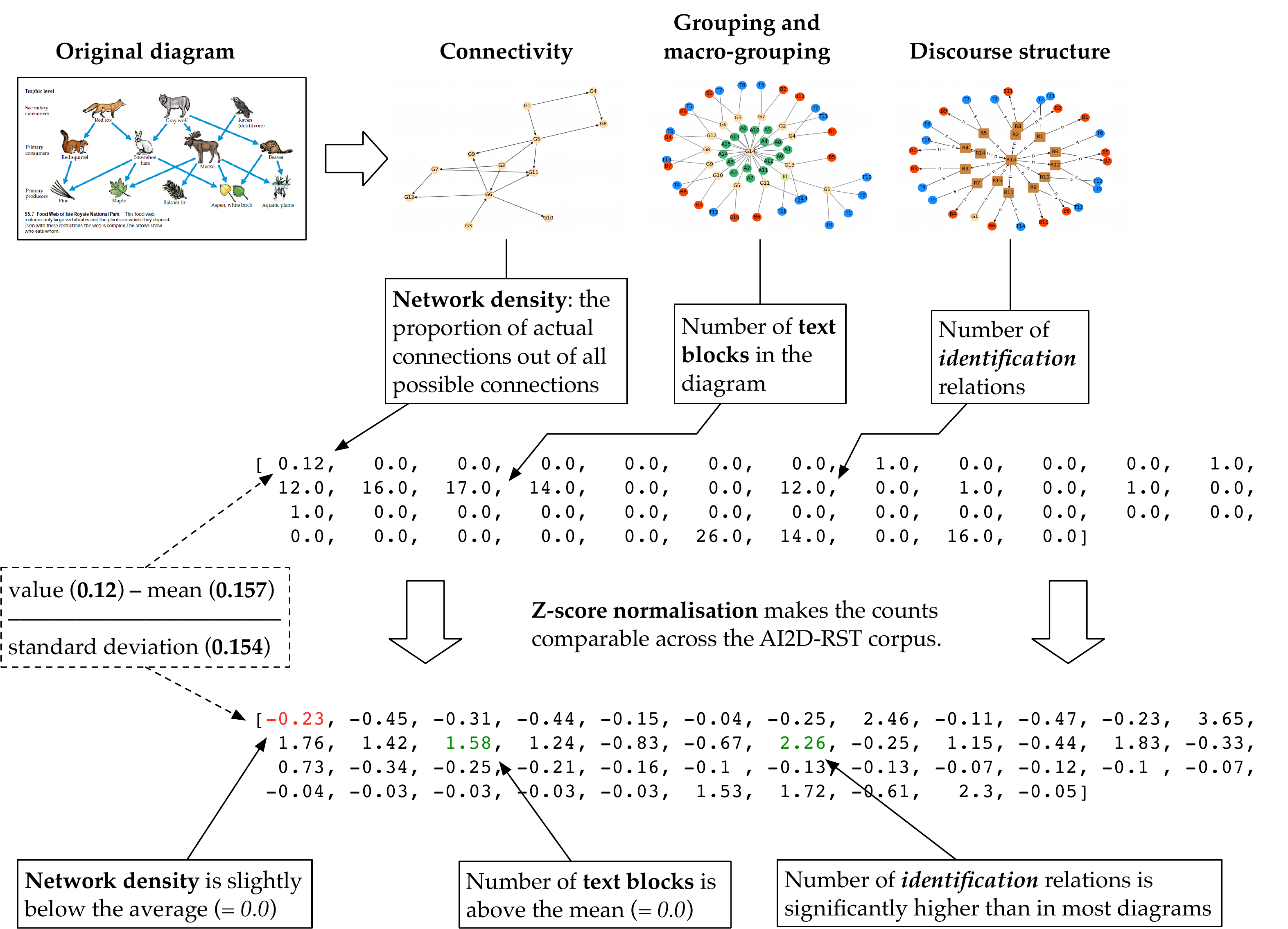}
    \caption{Extracting simple features from diagrams in the AI2D-RST corpus by counting the instances of different features across the annotation layers. The features are then normalised to make them comparable.}
    \label{fig:ai2d_method}
\end{figure}

In this section, we present a brief exploratory analysis of the AI2D-RST corpus. We begin with a rather straightforward approach illustrated in Figure \ref{fig:ai2d_method}, which makes minimal use of the graph-based representations by simply counting instances of diagram elements, macro-groups, rhetorical relations, nuclei and satellites, and types of connections in each diagram. Finally, we also calculate network density for the connectivity graph, which measures the proportion of actual edges present in the graph out of all possible edges. We concatenate these values into a 46-dimensional feature vector and use z-score normalization to scale the values of each dimension to have a mean of 0 and a standard deviation of 1. This provides each diagram in the AI2D-RST corpus with a normalised 46-dimensional feature vector that represents its the multimodal structure.

\begin{figure}[p]
    \centering
    \includegraphics[width=0.8\textwidth]{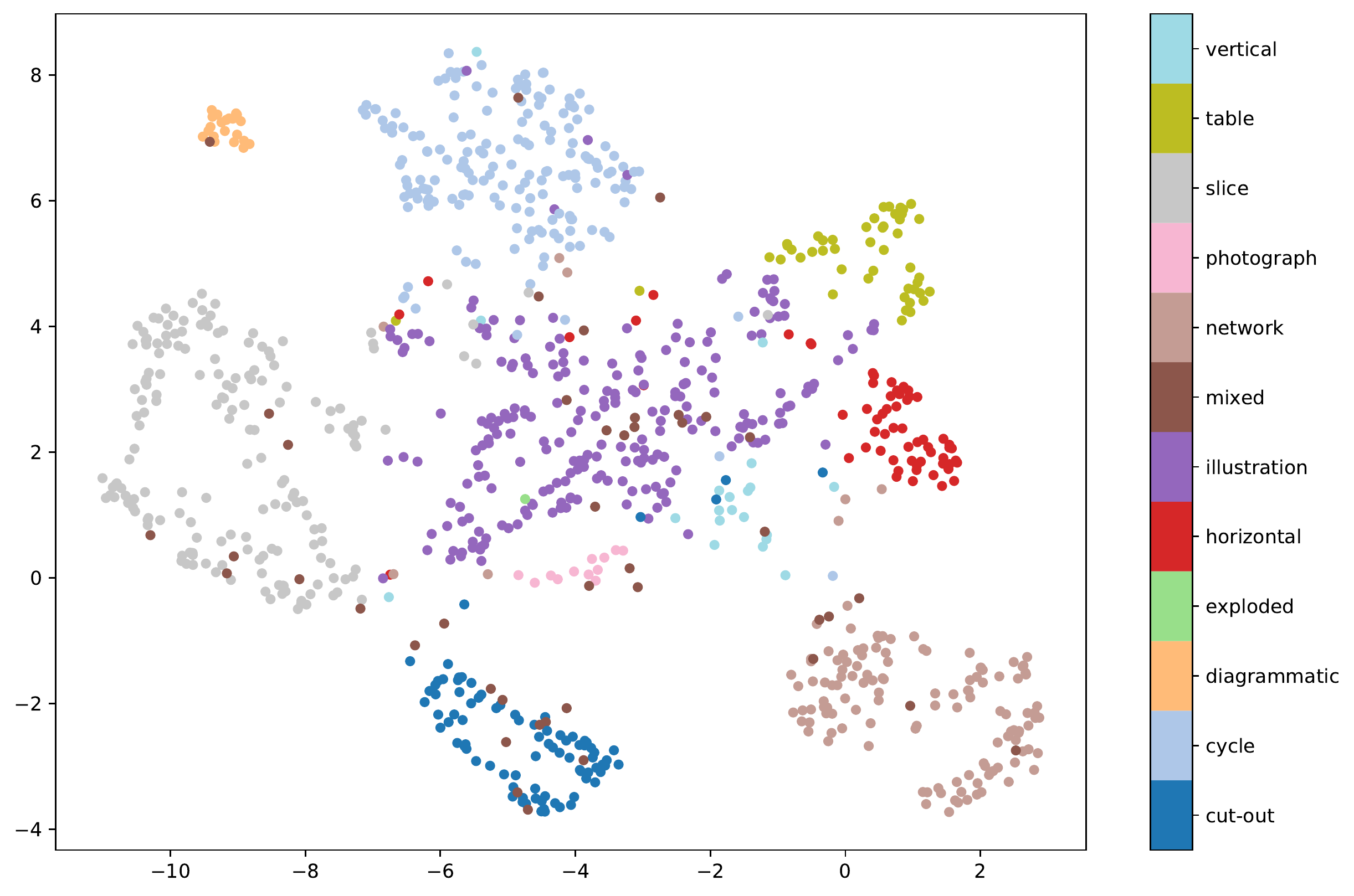}
    \caption{A visualization showing 2-dimensional UMAP embeddings learned from the 46-dimensional feature vectors extracted using the technique in Figure \ref{fig:ai2d_method}. Each point corresponds to a single diagram in the AI2D corpus, which are coloured according to their macro-groups.}
    \label{fig:umap}
\end{figure}

Figure \ref{fig:umap} shows a visualisation that uses the UMAP algorithm \citep{mcinnes2018-software} to reduce the 46-dimensional feature vectors to two dimensions for a visual exploration of the AI2D-RST corpus. When mapping points between high- and low-dimensional spaces, UMAP seeks to preserve both local and global structure of the points in the two spaces. In other words, points that are close to each other in the 46-dimensional space should be close to each other in the two-dimensional space, whereas points that are distant from each other in the 46-dimensional space should remain distant in the two-dimensional space as well. 

The UMAP embeddings in Figure \ref{fig:umap} show distinct clusters that correspond to specific macro-groups, such as cycles, cross-sections, cut-outs and networks, which illustrate the space of structural variation among the AI2D-RST diagrams. It should be noted, however, that the macro-grouping annotation is explicitly encoded into the 46-dimensional feature vector. This information is thus directly available to UMAP for learning the 2-dimensional embeddings, which the algorithm leverages when clustering points in the low-dimensional space.

Nevertheless, the visualisation in Figure \ref{fig:umap} can yield valuable insights into the structural variation among the AI2D-RST diagrams. Firstly, diagrams that feature several macro-groups (see Section \ref{sec:macrogrouping}) can be found within all major clusters, which suggests that even simple count-based features can capture structural distinctions in diagrams. The diagrams labelled as `mixed' are particularly interesting, as they may yield information on which macro-groups are readily combined with each other in the AI2D-RST corpus. The clusters for individual macro-groups, in turn, appear to capture variation within macro-groups, as exemplified by the clusters for networks and cross-sections, which seem to form two parts. Whether such formations within clusters reflect alternative structural configurations of expressive resources within specific diagram types warrants further analysis.

Secondly, diagrams that feature rigid layouts, such as tabular, horizontal and vertical macro-groups, are not only positioned close to each other, but also form a continuation of the cluster for illustrations. This is not surprising, as tabular, vertical and horizontal macro-groups are typically used to organise \emph{multiple instances} of visual depictions and their verbal descriptions for presentation, in which the local discourse structures are similar to \emph{individual} illustrations (for examples of local discourse structures, see Figure \ref{fig:rst_2185}). The clusters for cut-outs and cross-sections, in turn, are distinct from illustrations, which may be traced back to differences in their discourse structure. Whereas cut-outs and cross-sections typically use labels to pick out parts or regions of a visual depiction, illustrations use labels to identify the entire object. This distinction is captured by their discourse structure annotation.

Thirdly, the diagrammatic macro-group forms a tight cluster, which is clearly separate from other macro-groups. Although the sample size for this macro-group is fairly small ($N=22$), this is an interesting observation as the UMAP embeddings seem to capture a fundamental difference between the diagrammatic macro-group and other macro-groups in the corpus, which may be traced back to their discourse structure. The diagrammatic macro-group features schematic diagrams such as circuit diagrams, whose elements have \emph{fixed} meanings, as exemplified by standardized symbols for switches, connections, circuit breakers and the like.

Because their diagram elements have fixed meanings that do not need to be recovered discursively from their context of occurrence, schematic diagrams resist RST analysis. Put differently, there is no need for the viewer to resolve discourse relations between diagram elements, as all the information needed for making sense of the diagram is communicated using arrows and lines that signal connections between diagram elements with fixed meanings. Although these connections are captured by the AI2D-RST connectivity layer, this raises questions about the need to revise the AI2D-RST annotation schema, if it were to be extended to domains featuring many types of schematic diagrams, in order to draw out their differences.

This brief exploratory analysis has illustrated how the AI2D-RST corpus can be used to support empirical research on the multimodality of diagrams. As pointed out above, the features extracted from the corpus made minimal use of the properties of the graph-based representations (see Figure \ref{fig:ai2d_method}). The properties of graphs could be exploited to a much larger extent using algorithms such as graph neural networks \citep[see e.g.][]{wuetal2019}, which learn representations of graph-structured data by passing and receiving features between neighbouring nodes. Such methods could be particularly useful for learning representations of discourse structure in diagrams, allowing their computational representation to encode interactions between diagram elements. However, learning these representations directly from the data can be complicated by the relatively small number diagrams in AI2D-RST.

\section{Discussion}

Developing the AI2D-RST corpus showed that exploiting readily-available annotations can be used to increase the volume of richly-annotated multimodal corpora, but this comes at a cost, particularly for annotating their discourse structure. As explicated in \citet{hiippalabateman2020}, identifying the elementary discourse units required by RST and other discourse annotation frameworks is particularly complicated for diagrams, because the extent to which diagrams need to be decomposed to achieve a sufficient inventory of elementary discourse units varies from one diagram to another. In short, the level of detail needed for decomposition depends on the \emph{combination} of expressive resources and the discourse relations they participate in (see Section \ref{sec:resources}).

Because the AI2D layout segmentation does not provide this kind of discourse-driven decomposition at various levels of detail, the AI2D-RST annotation schema had to make compromises in the description of discourse structure. The example in Figure \ref{fig:rst_0} illustrates this issue aptly: the written labels are used to pick out parts of the illustration, and to achieve a maximally accurate RST analysis of the diagram, the illustration should be decomposed into its component parts. However, as the crowd-sourced annotators were not instructed to decompose visual expressive resources during layout segmentation, the elementary discourse units needed for a maximally coherent representation of discourse structure within RST are not available \citep[for a discussion of similar problems in annotating comics, see][]{batemanwildfeuer2014b}.

This shortcoming also carries implications for crowd-sourcing annotations for the diagrammatic mode in any domain. Because the discourse structure determines to what extent the diagram must be decomposed, defining crowd-sourcing tasks developed for the annotation of photographic images is unlikely to work for identifying the `building blocks'  of diagrams \citep[cf.][]{kovashkaetal2016}. Instead of defining semantic object classes (i.e. what the diagram element represents), these building blocks should correspond to expressive resources available to the diagrammatic mode, such as written language, arrows, lines and other diagrammatic elements. Crucially, these expressive resources must be complemented by sufficiently fine-grained descriptions of graphic expressive resources, such as line drawings, coloured illustrations, cut-outs, cross-sections and exploded views, and photographs, to name just a few examples. In short, pre-theoretical notions such as `language' and `image' \citep[cf.][]{bateman2014b} are not sufficiently fine-grained to capture the motivated use of distinctive graphic expressive resources in diagrams.

Although the development of AI2D-RST revealed various challenges discussed above, we argue that the corpus is still a valuable resource for studying how the diagrammatic mode is used in the domain of primary school natural sciences and beyond. In the study of multimodal discourse, the corpus could be used for investigating whether discourse relations between diagram elements are signalled visually using arrows and lines or spatially using layout \citep[cf.][]{watanabenagao1998}, thus complementing the linguistic research on signalling of discourse relations by \citet{dastaboada2018}. Such empirically-backed insights could be particularly valuable to educational research on the visual perception of diagrammatic representations and their role in constructing mental models \citep{tippett2016}. Another avenue of further research involves the automatic annotation of diagram corpora. The AI2D-RST corpus covers just over 20\% of the AI2D dataset, which raises the question whether the 1000 diagrams in AI2D-RST are sufficient for teaching algorithms to generate AI2D-RST representations for the remaining 3900 diagrams in AI2D.

\label{sec:discussion}

\section{Concluding remarks}
\label{sec:conclusion}

In this article we introduced AI2D-RST, a new multimodal corpus of 1000 English-language primary school science diagrams, which combines crowd-sourced and expert annotations to provide a rich description of their multimodal structure. The multi-layered, stand-off annotation schema developed for AI2D-RST accounts for (1) the visual grouping of diagram elements, (2) how their connections are signalled using arrows and lines, and (3) the discourse relations between diagram elements using Rhetorical Structure Theory. We measured agreement between five annotators: the results suggest that the annotation schema may be reliably applied to describe diagrams in the AI2D-RST corpus.

As our brief exploratory analysis of the AI2D-RST corpus showed, the combination of multiple annotation layers and graph-based representations can yield valuable insights into the multimodal structure of diagrams. As such, the corpus can support empirical research on diagrams as a mode of expression and their computational processing. In terms of methodology, developing the AI2D-RST corpus illustrated how crowd-sourcing low-level annotations and building expert descriptions on top of them can be used to increase the size of corpora in the field of multimodality research. Insights from linguistically-inspired multimodality research, in turn, can also inform the creation of resources for research on the computational processing and generation of diagrams. 


\bibliographystyle{agsm}
\bibliography{../../../bibliography}

\end{document}